\documentclass[conference]{IEEEtran}
\usepackage{blindtext, graphicx}
\usepackage{booktabs}
\usepackage{amsmath}
\usepackage{multirow}
\begin{document}

\title{A Machine-Learning Item Recommendation\\System for Video Games
}

\IEEEoverridecommandlockouts

\author{\IEEEauthorblockN{Paul Bertens,$^\dagger$ Anna Guitart,$^\dagger$\thanks{$^\dagger$These two authors contributed equally to the work.} Pei Pei Chen and \'{A}frica Peri\'a\~{n}ez}
 \IEEEauthorblockA{
 \textit{Yokozuna Data, Silicon Studio}\\
 1-21-3 Ebisu Shibuya-ku, Tokyo, Japan\\
 \{paul, anna, peipei, africa\}@yokozunadata.com}
 }

\maketitle

\begin{abstract}
Video-game players generate huge amounts of data, as everything they do within a game is recorded. In particular, among all the stored actions and behaviors, there is information on the in-game purchases of virtual products. Such information is of critical importance in modern free-to-play titles, where gamers can select or buy a profusion of items during the game in order to progress and fully enjoy their experience.
To try to maximize these kind of purchases, one can use a recommendation system so as to present players with items that might be interesting for them. Such systems can better achieve their goal by employing machine learning algorithms that are able to predict the rating of an item or product by a particular user. In this paper we evaluate and compare two of these algorithms, an ensemble-based model (extremely randomized trees) and a deep neural network, both of which are promising candidates for operational video-game recommender engines. 
Item recommenders can help developers improve the game. But, more importantly, it should be possible to integrate them into the game, so that
users automatically get personalized recommendations while playing. The presented models are not only able to meet this challenge, providing accurate predictions of the items that a particular player will find attractive, but also sufficiently fast and robust to be used in operational settings.  
\end{abstract}


\section{Introduction}

%
The aim of a recommender system is to provide suggestions to a set of users on items that might be interesting for them. Recommendation systems are commonly found in e-commerce \cite{sarwar2000analysis,linden2003amazon} (where users purchase goods like books, clothes or games online), usually implemented through collaborative filtering methods \cite{breese1998empirical}. These work by comparing similar items or similar users based on user ratings. If two users like the same items they are likely similar, and if two items are liked by the same users, those items are probably similar as well. However, as this method does not take into account the contents, new items cannot be recommended. Content-based recommenders can be used to overcome some of these issues by looking at the item in question and finding similarity between items based on inherit properties \cite{van2013deep}. A hybrid approach can also be taken, to combine e.g. collaborative information, content features and demographics \cite{gomez2016netflix}. A more detailed study into the current limitations and possible extensions of recommendation systems can be found in \cite{adomavicius2005toward}.

The integration of recommendation systems into video games is a relatively new area of research. Previous work has mostly focused on \emph{game} recommendation engines, which present players with suggestions on alternative titles based on the games they have already played \cite{anwar2017game, sifa2015large}. But it is also possible to use recommendation systems to increase player engagement in a game. In modern free-to-play games, users can buy a wide range of virtual items with real money (in-app purchases, IAPs). However, sometimes they can be overwhelmed by the number of items offered and the diversity of playstyles, and this can lead to an increase in the churn rate---as players start to find the contents too difficult and are unable to progress within the game. \emph{Item} recommendation systems can help prevent this problem by offering players a more direct route to the items that could be appealing or useful for them, thereby improving their purchasing and general in-game experience. This may ultimately result into increased revenue \cite{lehdonvirta2009virtual} by increasing player retention, IAPs and the conversion rate from free to paying users.

To achieve these goals, it is essential to recommend each player the right item---one that fits both their current state and their playing behavior---at the right time. And this is possible because (in contrast to other applications where very limited information is available) every action performed by a player within the game gets recorded. This offers a unique opportunity not only to obtain accurate predictions on the player's in-game behaviour (for example on when and at what level they will leave the game, see \cite{perianez2016churn} and \cite{bertens2017games}) but also to offer them personalized recommendations of items that are likely relevant to them.  

There are previous papers related to item recommendation systems. \cite{sifa2018controlling} introduces a recommendation system for the massively multiplayer online first-person shooter game \textit{Destiny}, where players get suggestions on those items that best fit their play style and might improve their performance. They apply similarity measures to global descriptors like total kill count or kill/death ratio. Clusters for the player ``base'' and ``cooldown'' stats were derived through $k$-means clustering, whereas archetypal analysis \cite{cutler1994archetypal, sifa2014archetypal} (which clusters by extreme values rather than centroids \cite{bauckhage2015k}) was used to find distinct playstyles. Similar analyses were done for the massively multiplayer online role-playing game \textit{Tera} and the multiplayer strategy game \textit{Battlefield 2: Bad Company 2} \cite{drachen2012guns} or the game \textit{Tomb Raider: Underworld} \cite{drachen2009player}. In all these cases, players were clustered by their playing behaviour; although no recommendation system was built, behavioral profiling via clustering may be very useful in offering recommendations based on similarity between users. 


However, unsupervised clustering methods remains a challenge. In particular, a significant amount of game-specific knowledge, is required to find adequate features that can separate players into the right number of clusters. 

\subsection{Aim}
While there are several approaches to the problem of developing recommendation systems, here we will explore a different avenue: our aim is to provide a method that predicts the next items a player will purchase, and use this information to recommend them other items. This approach differs from traditional methods as we explicitly use a predictive model. 

Such a model allows us to predict, both for new and existing users, the items they are likely to find most appealing based on their playing behaviour. 
Additionally, it must be robust for operational implementation, to be able to recommend game products automatically, in a variety of game genres, namely different game data distributions.

\section{Background}
\subsection{Extremely Randomized Trees}
Extremely randomized trees (ERTs) \cite{geurts2006extremely} extend the randomization of original random forest \cite{ho1998random,breiman2001random} algorithms by choosing the splitting points randomly instead of computing the ones that are more correlated with the output (which makes random forest an easy biased approach). ERTs are computationally efficient, reducing the variance of the model and preventing overfitting. However the bias can also be larger with this method when the randomization is increased above the optimal level, due to the decrease in the variance.


Breiman implementation of random forest builds an ensemble of decision trees, each of which is fit on a random subset of features \cite{breiman2001random}. This randomization in the feature selection, combined with the bagging of multiple decision trees, reduces the correlation between trees and increases the overall accuracy of the ensemble. 
    
One of the main advantages of ensemble models is that they are trivially parallelizable, either using multicore processors (as each tree could potentially be trained on a single core) or across multiple machines. This makes them more practical in operational settings, where training and inference have to be completed in a relatively short time, and thus better suited for developing a commercial recommendation system. 

\subsection{Deep Neural Networks}
Deep neural networks (DNNs) \cite{lecun2015deep} are artificial neural networks with multiple hidden layers. By using nonlinear activation functions (the functions that transform the output at each layer before passing it to the next), DNNs are able to learn highly nonlinear dynamics. Multiple iterations, i.e. epochs, are run to optimize the DNN during the learning process. Rectified linear units (ReLU) are among the most commonly used activation functions nowadays. DNNs that combine ReLU with dropout---a strategy consisting in randomly dropping out some of the units at each layer---have been shown to provide state-of-the-art accuracy in domains such as image classification \cite{krizhevsky2012imagenet} or speech recognition \cite{hinton2012deep}. Additionally, for sequential data, recurrent neural networks (RNN) or long short-term memory (LSTM) networks \cite{hochreiter1997long} have achieved similarly high accuracies in sequence prediction and language modeling. 


\section{Item Recommendation Model}

While RNNs and LSTM networks are able to learn temporal dependencies and eliminate the need for manual feature engineering, they also slow down the training significantly, as they have to learn the relevant features of the time series that lead to an increase in prediction accuracy. On the other hand, by manually calculating general statistics of the time-series data together with other descriptors one can efficiently create a single vector describing the player's behavior and use it in nontemporal models like DNNs or ensemble-based methods such as ERTs. 

These are the main challenges related to our approach:
\begin{itemize}
\item The model should be able to train and provide inference in production environments scaling to millions of users.
\item It should be trainable on mini-batches so that it fits in the memory (ensemble models usually work on the full dataset).
\item The time-series data needs to be converted into a single feature vector that accurately represents the player's behavioural patterns (as commented above, tree ensembles and DNNs use static feature vectors, not time series).
\item As players make multiple purchases over their lifetime in the game, we must extract their next purchase from multiple time points. Thus the training dataset may become huge if e.g. players remain in the game for several years.
\end{itemize}
The following sections elaborate on the dataset used and on the way the model was constructed to solve these challenges. 



\subsection{Dataset}

The data used in our analysis comes from the Japanese card-game \textit{Age Of Ishtaria}, developed by \textit{Silicon Studio}, and contains daily time-series data for each paying user within the period from 2014-09-24 to 2017-05-08 (totaling 33,488 players). It contains information on the number of purchases per item and total sales per item for each user. Players can purchase in-game currency with real money and use it to buy different card-packs (known as \emph{gacha}) containing a random set of cards that can be employed in the game. The data contains 8 different types of items and also has information on e.g. the player's daily level progression, playtime and lifetime.

\subsection{Feature calculation}

To convert our time-series into a single static vector we calculate general statistics over the full time-series data for each of the temporal features (e.g. daily playtime or sales). The process is as follows: First we compute the derivative of the time series in order to get its variations (for instance, if we are tracking total level, the derivative gives us the number of level-ups per day). Then we calculate the mean/variance/skew/kurtosis/maximum over the time series for each of the temporal features. Additionally, to capture behavioral changes of the player between the beginning and end of their lifetime, we also compute the distance for all temporal features over the first and last days in which they logged in. Finally, all these features get concatenated into our final feature vector. By using such a method, the feature calculation can be generalized to any type of temporal data.

\subsection{Sampling to handle multi-label outputs}

Players usually make multiple purchases, which means we can have multiple prediction targets (multiple labels) per user. One way of dealing with this is taking some subsample until time $t$ from each player's time series and then find their next purchase after $t$. This results in a single label we can train on, and allows us to take multiple subsamples to enlarge our training set. Since players could be playing for several years and have hundreds or even thousands of days of playing activity, by using subsampling we can generate different training samples for each player, increasing our effective training dataset and reducing overfitting. 

\subsection{Scalability using minibatches}

Additionally, the model should be able to scale to millions of players; however, if we generate very large feature vectors (with thousands of features) and sample multiple labels per user, we could end up with datasets with over a billion samples (a thousand samples per user). An efficient way of coping with such huge data sets is to train an ensemble model on subsamples of the total set. Hence, we can train a small subset of trees ({\raise.17ex\hbox{$\scriptstyle\mathtt{\sim}$}}20) on a small sample of a few thousand users and generate the labels directly during training, so that we do not need to store all samples. The final ensemble is formed by combining many such subsets of trees, where each tree was trained on different features, different samples, and different target labels, producing an extremely robust model.

\subsection{Model Specification}

\subsubsection{Output}
For each player and item, we generate the probability that they will buy that item on their next purchase day. As the model is trained over all players, once players are in a similar state the model can learn to predict and recommend the right item at the right time for each individual player.  

\subsubsection{Input}
We take the full time-series patterns for each user to convert them into a single vector that represents their playing behavior. This conversion is done for all users in a single mini-batch. Multiple mini-batches are generated per epoch (one epoch goes over the entire dataset), and the model is trained on each of these batches. 

\subsubsection{Parameters}
The ERT model was trained on subsets of 20 trees for 30 iterations, resulting in a total ensemble size of 600 trees. Each iteration was performed on a subset of {\raise.17ex\hbox{$\scriptstyle\mathtt{\sim}$}}10k users, which means that a full single epoch was completed after 3 iterations (as the total set has 33,488 players), therefore we had 10 epochs.

For the DNN model, we used two hidden layers of 2048 units and set a dropout probability of 0.5. Additionally, as there were many correlating features, dropout was also applied to the input layer. By randomly dropping some inputs, we reduce overfitting on single features, thereby increasing the robustness of the model. (Recall this was achieved by random subsampling of features in the ERT model.) The network was trained for 30 iterations as well, but each iteration was repeated 5 times, resulting in a total of 50 epochs. Both DNN and ERT are trained on the same data.

\section{Model Evaluation}
In order to evaluate the effectiveness of the proposed model, we study the prediction accuracy within an upcoming time window. Predictions are made at a time point $t$ and evaluated at time $t+50$ (where $t$ is measured in days). The training was performed using data up to 2017-03-19, and predictions were verified in the window from 2017-03-20 to 2017-05-08. 

Several measures are calculated:

1) \textit{isOnNextPurchaseDate}: Checks whether the predicted item was actually acquired by the player throughout their next purchase day (our training objective). 

2) \textit{isNextPurchase}: Checks whether the item that was predicted to be purchased by a certain player  was actually acquired by the player on their very next purchase.

3) \textit{isWithinWindow}: Checks whether the predicted item was actually acquired by the player at some point within the time window considered (between $t+1$ and $t+50$).

For all three measures, the accuracy for the top (\textit{predictedMax}), top 2 (\textit{withinTop2}) and top 3 (\textit{withinTop3}) predicted items is calculated, i.e.\ we check whether the player actually purchased the item that had the highest probability, any of the two items with the two highest probabilities or any of the three items with the three highest probabilities, as per the prediction. 


\section{Results}

\begin{figure}[h!]
  \centering
   \includegraphics[height=0.3\textwidth]{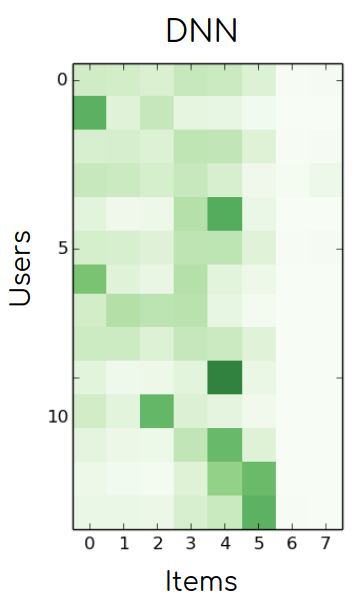}  
  	\hspace{0.5cm}
   \includegraphics[height=0.3\textwidth]{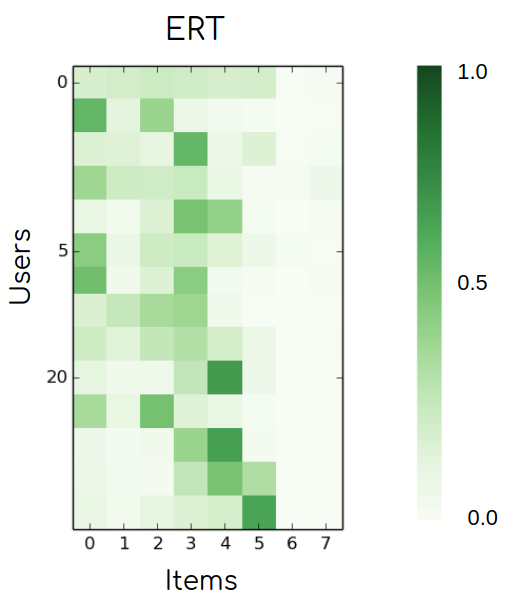} 
   \caption{Predicted probability, for a sample of players and a series of items, that the item will be bought by the player on their next purchase, using  the DNN (left) and ERT models (right). (Darker colors correspond to higher probabilities.) }
\label{item-user-prediction-matrix}
\end{figure}



Figure \ref{item-user-prediction-matrix} shows the predictions for a subset of users. The DNN (left panel) and ERT (right panel) results exhibit similar patterns (with only slight variations). We see that different users have different purchase probabilities for each item, which shows that the models are capable of providing personalized predictions for each player based on their playing behaviour. 

The accuracy results for both models can be found in Table \ref{DNN-table}. When considering the top 2 and top 3 predictions, both models present similar accuracies, but the ERT is slightly better at identifying the item with the highest probability of being acquired on the next purchase, for all three measures.  

\begin{table}[tbh!]
\centering
\caption[Accuracy results for the next-purchase prediction in the DNN and ERT models]{Accuracy results for the next-purchase\\prediction in the DNN and ERT models}
\label{DNN-table}
\begin{tabular}{@{}lccc@{}}
\toprule
\multicolumn{1}{c}{(DNN)} & predictedMax  & withinTop2    & withinTop3    \\ \midrule
isOnNextPurchaseDate            & 44\% & 68\% & 81\% \\
isNextPurchase     				 & 34\% & 59\% & 74\% \\
isWithinWindow            & 69\% & 85\% & 90\% \\ \bottomrule

\toprule
\multicolumn{1}{c}{(ERT)} & predictedMax  & withinTop2    & withinTop3    \\ \midrule
isOnNextPurchaseDate      & \textbf{47\%} & 68\% & 81\% \\
isNextPurchase            & \textbf{37\%} & 59\% & 74\% \\
isWithinWindow            & \textbf{71\%} & 85\% & 91\% \\ \bottomrule
\end{tabular}
\end{table}




\section{Discussion}
An item recommendation system for games is essential to provide players with individual rewards or incentives to increase engagement, to maximize in-app purchases and to increase cross-selling and up-selling. We have presented two models to predict which items players will be more attracted to buy in their next purchases. The results show that the predicting performance of the DNN and ERT  is similar. However the ERT model yields slightly better results (as shown in Table~\ref{DNN-table}) and also scales up more easily in a production environment.


While predictions were made only for a small set of items, the model is trivially extendable to run on hundreds of items, and can be used both for items purchased with real money and for in-game virtual purchases. Future works in this direction will include an evaluation of the recommendation system in terms of total game sales for live video-games.


\section*{Acknowledgements}
We thank Javier Grande for his careful review of the manuscript and Ana Fern\'andez for her support.

\bibliographystyle{abbrv}
\bibliography{main.bib}	

\end{document}